\documentclass{article}

\usepackage[final,nonatbib]{nips_2016}

\usepackage[utf8]{inputenc} 
\usepackage[T1]{fontenc}    
\usepackage{hyperref}       
\usepackage{url}            
\usepackage{booktabs}       
\usepackage{amsfonts}       
\usepackage{nicefrac}       
\usepackage{microtype}      
\usepackage{graphicx}

\usepackage[caption=false,font=footnotesize]{subfig}
\usepackage[justification=centering]{caption}

\usepackage{xcolor}
\newcommand{\re}{\textcolor[rgb]{0.2,0.6,0.2}}
\newcommand{\syn}{\textcolor[rgb]{0.9,0.2,0.2}}

\title{Towards Adversarial Retinal Image Synthesis}

\author{P. Costa \hspace{1cm} A. Galdran \hspace{1cm} M.I. Meyer   \\[1mm]
  INESC TEC Porto, Instituto de Engenharia de Sistemas e Computadores - Tecnologia e Ciência \\
  Porto, E-4200, Portugal \\
  \texttt{\{pedro.costa, adrian.galdran, maria.i.meyer\}@inesc.pt} \\
   \AND  
  M. D. Abràmoff  \\
   Stephen A Wynn Institute for Vision Research, \\University of Iowa Hospital and Clinics\\
   Iowa City, IA 52242, USA \\
   \texttt{michael-abramoff@uiowa.edu} 
  \And 
  M. Niemeijer \\
   IDx LLC \\
   Iowa City, IA 52246, USA \\
   \texttt{niemeijer@eyediagnosis.net} \\
   \And
  A.M. Mendonça \hspace{1cm}  A. Campilho \\
  Faculty of Engineering, University of Porto \\
  Porto, E-4200-464, Portugal \\
  \texttt{\{amendon, campilho\}@fe.up.pt} \\
}

\begin{document}

\maketitle

\begin{abstract}
Synthesizing images of the eye fundus is a challenging task that has been previously approached by formulating complex models of the anatomy of the eye. New images can then be generated by sampling a suitable parameter space. In this work, we propose a method that learns to synthesize eye fundus images directly from data. For that, we pair true eye fundus images with their respective vessel trees, by means of a vessel segmentation technique. 
These pairs are then used to learn a mapping from a binary vessel tree to a new retinal image. For this purpose, we use a recent image-to-image translation technique, based on the idea of adversarial learning. Experimental results show that the original and the generated images are visually different in terms of their global appearance, in spite of sharing the same vessel tree. 
Additionally, a quantitative quality analysis of the synthetic retinal images confirms that the produced images retain a high proportion of the true image set quality.
\end{abstract}

\section{Introduction}\label{sec_intro}

Modern machine learning methods require large amounts of data to be trained. This data is rarely available in the field of medical image analysis, since obtaining clinical annotations is often a costly process. Therefore, the possibility of synthetically generating medical visual data is greatly appealing, and has been explored for years. However, the realistic generation of high-quality medical imagery still remains a complex unsolved challenge for current computer vision methods.

Early methods for medical image generation consisted of digital phantoms, following simplified mathematical models of human anatomy \cite{collins_design_1998}. These models slowly evolved to more complex techniques, able to reliably model relevant aspects of the different acquisition devices. When combined with anatomical and physiological information arising from expert medical knowledge, realistic images can be produced \cite{fiorini_automatic_2014}. These are useful to validate image analysis techniques \cite{hodneland_physical_2016}, for medical training \cite{liu_simulation_2010}, therapy planning \cite{cai_integrated_2014}, and a wide range of applications.

\begin{figure}[t]
\centering
\includegraphics[width = \textwidth]{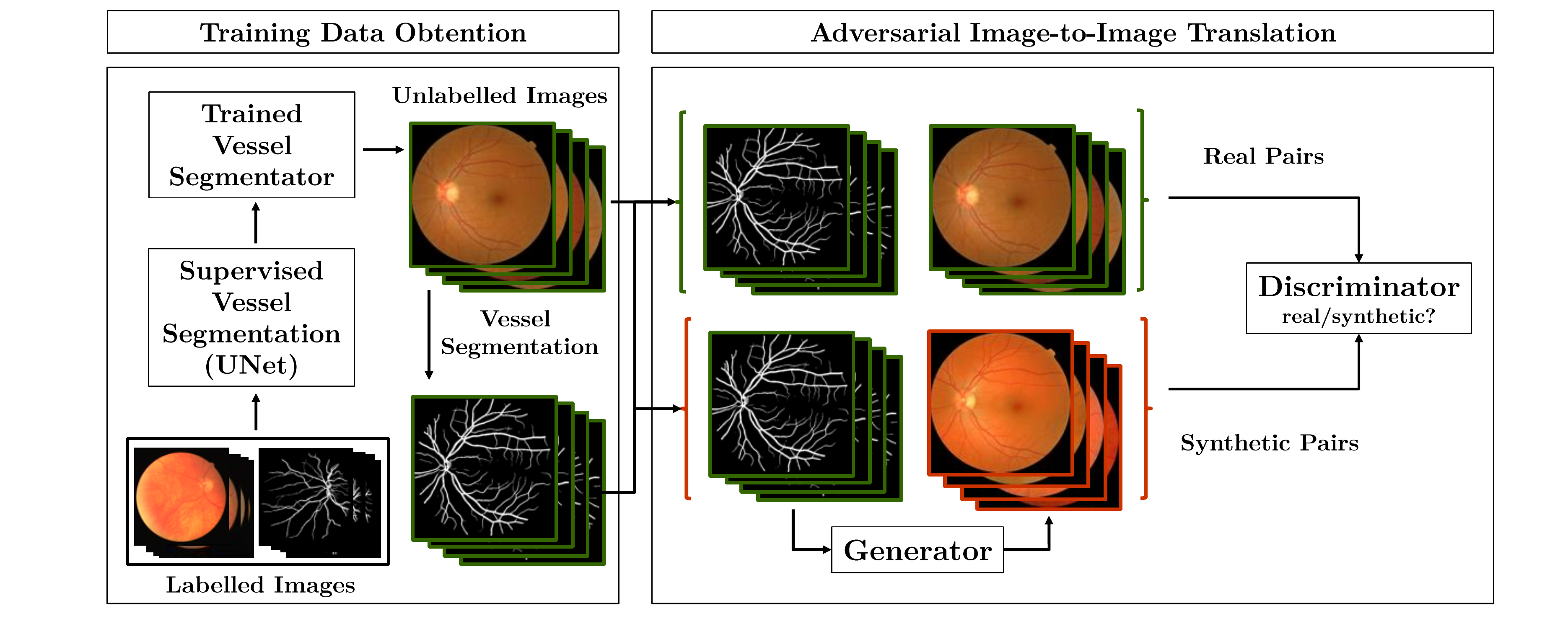}
\caption{Overview of the proposed retinal image generation method.}
\label{fig_overall}
\end{figure}
However, the traditional top-down approach of observing the available data and formulating mathematical models that explain it (\textit{image simulation}) implies modeling complex natural laws by unavoidably simplifying assumptions. More recently, a new paradigm has arisen in the field of medical image generation, exploiting the bottom-up approach of directly learning from the data the relevant information. This is achieved with machine learning systems able to automatically learn the inner variability on a large training dataset \cite{tulder_why_2015}. Once trained, the same system can be sampled to output a new but plausible image (\textit{image synthesis}). 

In the general computer vision field, the synthesis of natural images has recently experimented a dramatic progress, based on the general idea of adversarial learning \cite{goodfellow_generative_2014}. In this context, a generator component synthesizes images from random noise, and an auxiliary discriminator system trained on real data is assigned the task of discerning whether the generated data is real or not. In the training process, the generator is expected to learn to produce images that pose an increasingly more difficult classification problem for the discriminator. 

Although adversarial techniques have achieved a great success in the generation of natural images, their application to medical imaging is still incipient. This is partially due to the lack of large amounts of training data, and partially to the difficulty of finely controlling the output of the adversarial generator. In this work, we propose to apply the adversarial learning framework to retinal images. Notably, instead of generating images from scratch, we propose to generate new plausible images from binary retinal vessel trees. Therefore, the task of the generator remains achievable, as it only needs to learn how to generate part of the retinal content, such as the optical disk, or the texture of the background (Figure \ref{fig_overall}).

The remaining of this work is organized as follows: we first describe a recent generative adversarial framework \cite{isola_image--image_2016} that can be employed on pairs of vessel trees and retinal images to learn how to map the former to the latter. Then, we briefly review U-Net, a Deep Convolutional Neural Network architecture designed for image segmentation, which allows us to generate pairs of retinal images and corresponding binary vessel trees. This model provides us with a dataset of vessel trees and corresponding retinal images that we then use to train an adversarial model, producing new good-quality retinal images out of a new vessel tree. 
Finally, the quality of the generated images is evaluated qualitatively and quantitatively, and a description of potential future research directions is presented.

\section{Adversarial Retinal Image Synthesis}\label{sec_aris}

\subsection{Adversarial Translation from Vessel Trees to Retinal Images}\label{subsec_from_bin}

Image-to-image translation is a relatively recent computer vision task in which the goal is to learn a mapping $G$, called \emph{Generator}, from an image $x$ into another representation $y$ \cite{isola_image--image_2016}. Once the model has been trained, it is able to predict the most likely representation $G(x_{new})$ for a previously unseen image $x_{new}$. 

However, for many problems a single input image can correspond to many different correct representations. If we consider the mapping $G$ between a retinal vessel tree $v$ and a corresponding retinal fundus image $r$, variations in color or illumination may produce many acceptable retinal images that correspond to the same vessel tree, i.e. $G(v) = \{r_1, r_2,\ldots, r_n\}$. Directly related to this is the choice of the objective function to be minimized while learning $G$, which turns out to be critical. Training a model to naively minimize the $L2$ distance between $G(v_i)$ and $r_i$ for a collection of training pairs given by $\{(r_1,v_1), \ldots, (r_n,v_n)\}$ is known to produce low-quality results with lack of detail \cite{lotter_pgn_2015}, due to the model selecting an average of many equally valid representations. 

\begin{figure}[t]
\centering
\includegraphics[width = 0.8\textwidth]{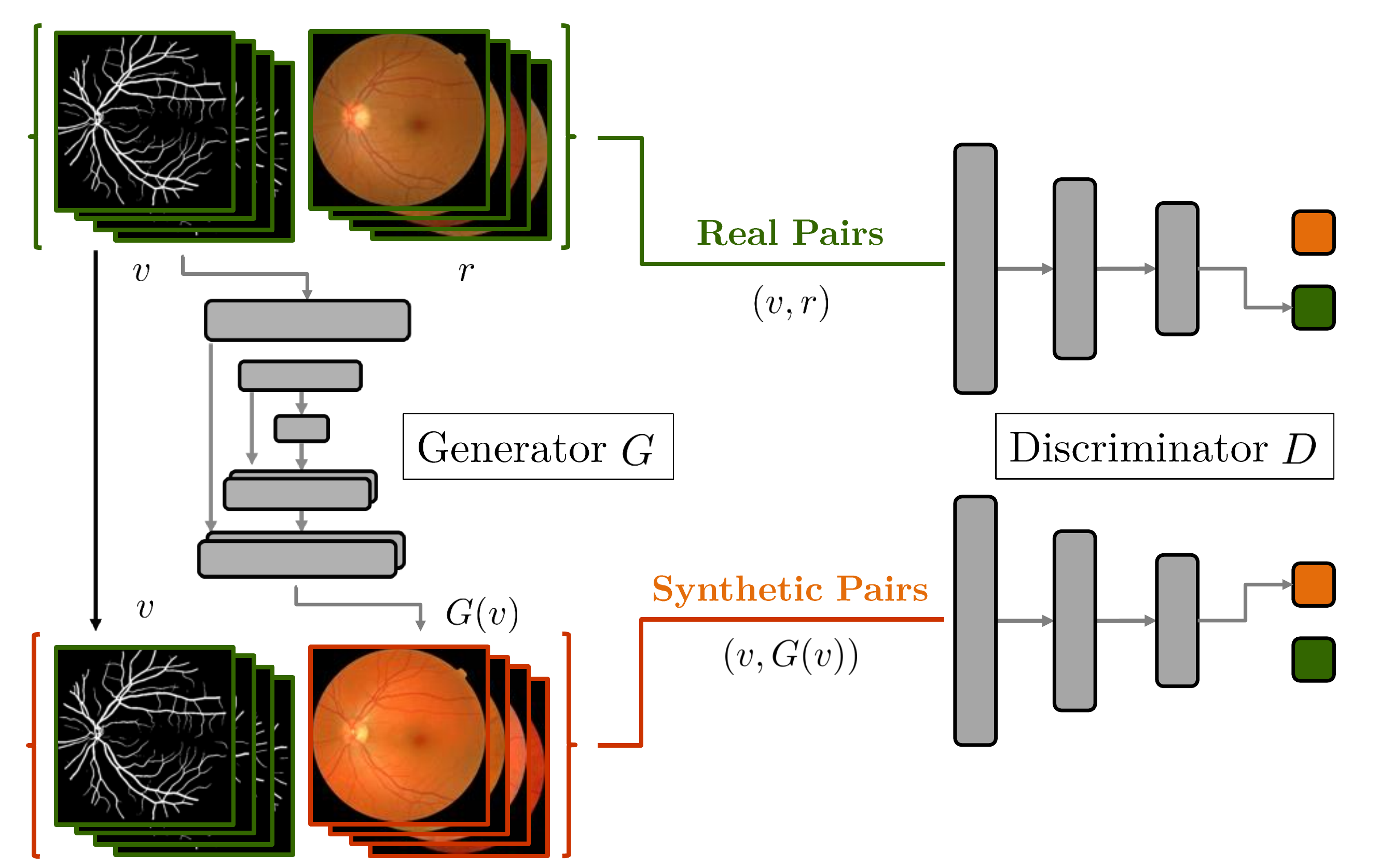}
\caption{Overview of the generative model mapping vessel trees to retinal images.}
\label{fig_pix2pix}
\end{figure}

Instead of explicitly defining a particular loss function for each task, it is possible to employ Generative Adversarial Networks to implicitly build a more appropriate loss \cite{isola_image--image_2016}. In this case, the learning process attempts to maximize the misclassification error of a neural network (called \emph{Discriminator}, $D$) that is trained jointly with $G$, but with the goal of discriminating between real and generated images. This way, not only $G$ but also the loss are progressively learned from examples, and adapt to each other: while $G$ tries to generate increasingly more plausible representations $G(v_i)$ that can deceive $D$, $D$ becomes better at its task, thereby improving the ability of $G$ to generate high-quality samples. Specifically, the adversarial loss is defined by:
\begin{equation}
\mathcal{L}_{adv}(G, D) = \mathbb{E}_{v,r \sim p_{data}(v, r)} [log D(v, r)] + \mathbb{E}_{v \sim p_{data}(v)} [ log(1 - D(v, G(v))) ],
\end{equation}
where $\mathbb{E}_{v,r \sim p_{data}}$ represents the expectation of the log-likelihood of the pair $(v,r)$ being sampled from the underlying probability distribution of real pairs $p_{data}(v,r)$, while $p_{data}(v)$ corresponds to the distribution of real vessel trees. An overview of this process is shown in Figure \ref{fig_pix2pix}.

To generate realistic retinal images from binary vessel trees, we follow recent ideas from \cite{shrivastava_su_learning_2016, isola_image--image_2016}, which propose to combine the adversarial loss with a global $L1$ loss to produce sharper results. Thus, the loss function to optimize becomes:
\begin{equation}\label{loss}
\mathcal{L}(G, D) = \mathcal{L}_{adv}(G, D) + \lambda \mathbb{E}_{v,r \sim p_{data}(v, r)} \left(||r - G(v)||_1\right),
\end{equation}
where $\lambda$ balances the contribution of the two losses. The goal of the learning process is thus to find an equilibrium of this expression. The discriminator $D$ attempts to maximize eq. (\ref{loss}) by classifying each $N\times N$ patch of a retinal image, deciding if it comes from a real or synthetic image, while the generator aims at minimizing it. The $L1$ loss controls low-frequency information in images generated by $G$ in order to produce globally consistent results, while the adversarial loss promotes sharp results. Once $G$ is trained, it is able to produce a realistic retinal image from a new binary vessel tree.

\subsection{Obtaining Training Data}\label{sec_vessel_seg}
The model described above requires training data in the form of pairs of binary retinal vessel trees and corresponding retinal images. Since such a large scale manually annotated database is not available, we apply a state-of-the-art retinal vessel segmentation algorithm to obtain enough data for the model to learn the mapping from vessel trees to retinal images. There exist a large number of methods capable of providing reliable retinal vessel segmentations. Here we employ a supervised method based on Convolutional Neural Networks (CNNs), namely the U-Net architecture, first proposed in \cite{ronneberger_u-net:_2015} for the segmentation of biomedical images. This technique is an extension of the idea of Fully-Convolutional Networks, introduced in \cite{shelhamer_fcn_2015}, adapted to be trained with a low number of images and produce more precise segmentations.

The architecture of the U-Net consists of a downsampling and an upsampling block. The first half of the network follows a typical CNN architecture, with stacked convolutional layers of stride two and Rectified Linear Unit (ReLU) activations. 
The second part of the architecture upsamples the input input feature map symmetrically to the downsampling path.
The feature map of the last layer of the downsampling path is upsampled so that it has the same dimension of the second last layer. The result is concatenated with the feature map of the corresponding layer in the downsampling path, and this new feature map undergoes convolution and activation. This is repeated until the upsampling path layers reach the same dimensions as the first layer of the the network. 

\begin{figure}[t]
	\centering
	\includegraphics[width = \textwidth]{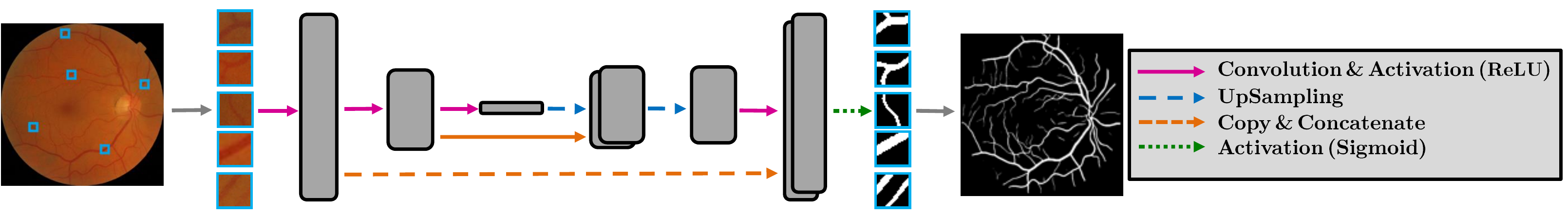}
	\caption{Overview of the U-Net architecture. Each box corresponds to a multi-channel feature map.}
	\label{fig: fig_unet}
\end{figure}

The final layer is a convolution followed by a sigmoid activation in order to map each feature vector into vessel/non-vessel classes. The concatenation operation allows for very precise spatial localization, while preserving the coarse-level features learned during the downsampling path. A representation of this architecture as used in the present work is depicted in Figure \ref{fig: fig_unet}.

\subsection{Implementation}
For the purpose of retinal vessel segmentation, the DRIVE database \cite{staal:2004-855} was used to train the method described in the previous Section. The images and the ground truth annotations were divided into overlapping patches of $64\times64$ pixels and fed randomly to the U-Net, with 10\% of the patches being used for validation. The network was trained using the Adam optimizer \cite{kingma:adam_2014} and binary crossentropy as the loss function.

Retinal vessel segmentation using the U-Net was evaluated on DRIVE's test set, achieving a $0.9755$ AUC, aligned with state-of-the-art results \cite{Liskowski_2016}. The optimal binarization threshold maximizing the Youden index \cite{Youden_1950} was selected. Messidor \cite{decenciere_feedback_2014} images were cropped, in order to only display the field of view, and downscaled to $512\times512$. Then, the segmentation method was applied to these images.
Messidor contains $1200$ images annotated with the corresponding diabetic retinopathy grade, and displays more color and texture variability than DRIVE's $20$ training images. Due to the U-Net being trained and tested in different datasets, some of the produced segmentations were not entirely correct. This may be related to DRIVE only containing $7$ examples of images with signs of mild diabetic retinopathy (grade 1). For this reason, we decided to retain only pairs of images and vessel trees in which the corresponding image had grade 0, 1, and 2.

The final dataset collected for training our adversarial model consisted of $946$ Messidor image pairs. This dataset was further randomly divided into training ($614$ pairs), validation ($155$ pairs) and test ($177$ pairs) sets. Regarding image resolution, the original model in \cite{isola_image--image_2016} used pairs of $256\times256$ images, with a U-Net-like generator $G$. We modified the architecture to handle $512\times512$ pairs, which is closer to the resolution of DRIVE images. For that, we added one layer to the downsampling part and another to the upsampling part of $G$. The discriminator $D$ classifies $16\times16$ overlapping patches of size $63\times63$. The implementation was developed in Python using Keras\footnote{Code to reproduce our results is available at \url{https://github.com/costapt/vess2ret}} \cite{chollet_keras_2015}. The learning process starts by training $D$ with real $(v, r)$ and generated pairs $(v, G(v))$. Then, $G$ is trained with real $(v, r)$ pairs. This process is repeated iteratively until the losses of $D$ and $G$ stabilize. 

\section{Experimental Evaluation}\label{sec_evaluation}
\begin{figure}[t]
\centering
\includegraphics[width = \textwidth]{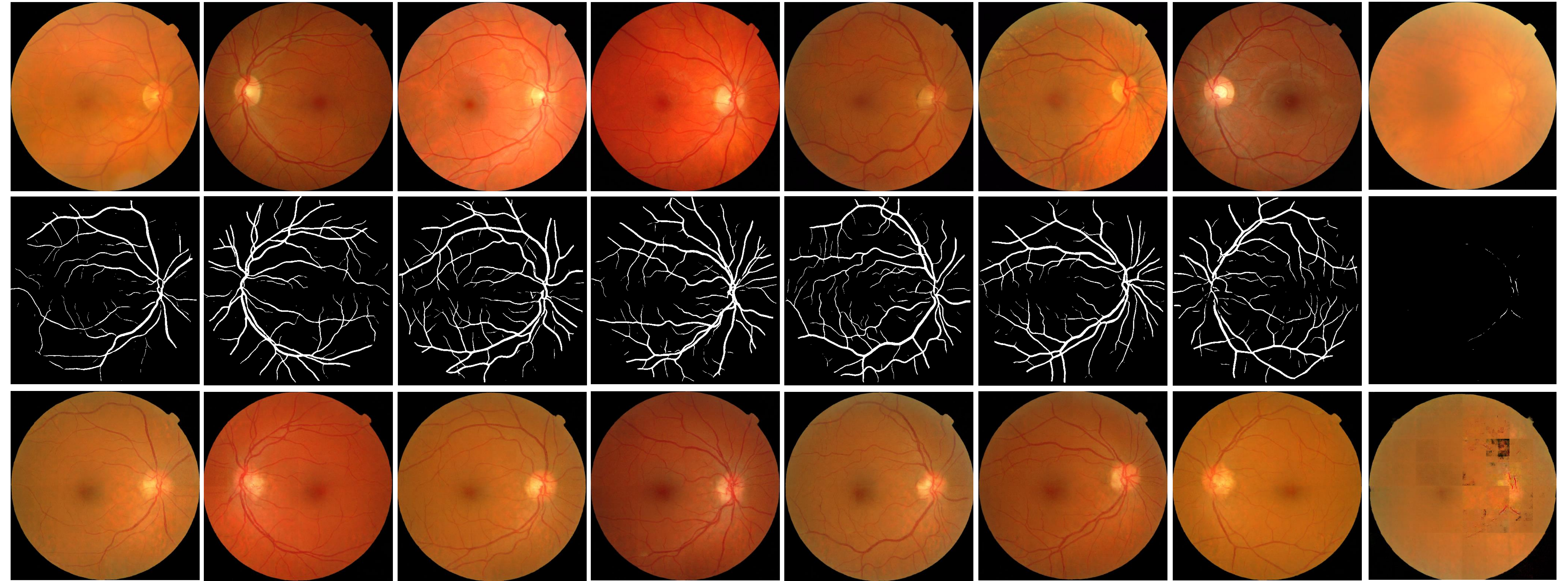}
\caption{Results of our model. First row: True retinal images from our test set, not used during training. Second row: Vessel trees obtained after segmenting images in the first row. Third row: Corresponding retinal images generated by our model. All images have $512\times512$ resolution.}
\label{fig_evaluation_1}
\end{figure}

For a subjective visual evaluation of the images generated by our model, we show in Figure \ref{fig_evaluation_1} some results. 
The first row depicts a random sample of real images extracted from the held-out test set, which was not used during training. 
The second row shows vessel trees segmented from those images with the method outlined in Section \ref{sec_vessel_seg}, and the bottom row shows the synthetic retinal images produced by the proposed technique. 
We see that the original and the generated images share some global geometric characteristics. This is natural, since they approximately share the same vascular structure. However, the synthetic images have markedly different high-level visual features, such as the color and tone of the image, or the illumination. This information was extracted by our model from the training set, and effectively applied to the input vessel trees in order to produce realistic retinal images.

The first seven columns of Figure \ref{fig_evaluation_1} show results in which the model behaved as expected: the vessel trees retrieved from the images in the first row were approximately correct, and provided sufficient information for the generator to create new consistent information in the synthetic image, shown in the last row. The last column in Figure \ref{fig_evaluation_1} shows a failure case of the proposed technique. Therein, the segmentation technique described in Section \ref{sec_vessel_seg} failed to produce a meaningful vessel network out of the original image. This is probably due to the high degree of defocus that the input image had. In this situation, the binary vessel tree supplied to the generator contained too few information, leading to the appearance of spurious artifacts and chromatic noise in the synthetic image. Fortunately, the amount of cases in which this happens was relatively low: out of our test set of $177$ images, $6$ were found to suffer from artifacts.

Objective image quality verification is known to be a hard challenge when no reference is available \cite{wang_why_2002}. In addition, for generative models it has been recently observed that specialized evaluation should be performed for each problem \cite{theis_note_2016}. In our case, to achieve a meaningful objective quantitative evaluation of the quality of the generated images, we apply two different retinal image quality metrics, namely the $Q_v$ score, proposed in \cite{koler_2013}, and the Image Structure Clustering (ISC) metric \cite{niemeijer_image_2006}. Both metrics have been employed previously to assess the quality of retinal images. While the $Q_v$ score focuses more on the assessment of contrast around vessel pixels, the ISC metric performs a more global evaluation. Thus, together they provide an appropriate mechanism to quantitatively evaluate the correctness of a synthetically generated retinal image. 

It is worth noting that in cases where artifacts and distortions were generated due to the undercomplete vessel network problem explained above, the ISC metric tended to artificially rise the quality of the synthetic image, as compared to the real one. Due to this, synthetic images containing this class of degradations were manually identified and removed from the ISC metric analysis below, together with their real counterparts. A more detailed discussion of both of the employed retinal image quality metrics, and their behavior when distorted images where supplied to them is provided in appendix \ref{sec_quality_disc}, together with supplementary results generated by the proposed technique.


The ISC score was computed on a reduced test set of 171 images (after removing the $6$ images with visual artifacts), while the $Q_v$ score was computed on all the $177$ images. The statistical analysis performed on both quality score distributions showed that both were normal according to the Kolmogorov-Smirnov test. The resulting data was therefore expressed as mean $\pm$ standard deviation, and compared with the paired Student's t-test. All $p$-values were two-tailed and $p< 0.05$ was considered significant. Statistical analyses were performed using GraphPad Prism 7 (Graphpad Software Inc.) software. Results obtained with this methodology are shown in Table \ref{qv_measure}. 

\begin{table}[t]
\centering

\begin{tabular}{lcccc}
\toprule[1.0pt]
\addlinespace[0.5em]
\hspace{0.35cm}	& \hspace{0.1cm} Mean $ISC$ score	\hspace{0.1cm}  	& \hspace{0.1cm} Std. dev. \hspace{0.1cm} & \hspace{0.1cm} Mean $Q_v$ score \hspace{0.1cm} & \hspace{0.1cm} Mean $Q_v$ score	\hspace{0.1cm} \\
\midrule[1.25pt]\addlinespace[0.5em]
Real Images 					  & $0.9872$ & $0.0468$ & $\mathbf{0.1254}$ & $\mathbf{0.0340}$\\
\addlinespace[0.5em]
Synthetic Images  					  & $0.9889$ & $0.0398$  & $\mathbf{0.1047}$ & $\mathbf{0.0136}$\\
\addlinespace[0.25em]
\bottomrule[1.5pt]\addlinespace[0.5em]
\end{tabular}
\caption{Result of computing the ISC and $Q_v$ quality measures on real/synthetic images. \\Statistically significant results are shown in bold.}
\label{qv_measure}
\end{table} 

In the case of the ISC metric, the synthetic images produced a slightly higher quality score, with the difference between them not statistically significant ($p = 0.2188$). For the $Q_v$ score, the real images were considered to be of better quality with regard to their synthetic counterparts, the difference being statistically significant ($p < 0.05$). However, it should be considered that the $Q_v$ score consists of an anisotropy measure weighted by the values of a simple vessel detector (see Appendix \ref{sec_quality_m}). In this case, it can be expected that image regions around the vessels of a synthetic image won't probably be of a better quality than the original ones. On the other hand, results on the ISC metric, which has a more global nature, point to a similar quality in the real and synthetic images, which agrees with the subjective visual quality found in the produced images, see Appendix \ref{supp_res}.

\section{Conclusions and Future Work}\label{sec_disc_fw}
The above visual and quantitative results demonstrate the feasibility of learning to synthesize new retinal images from a dataset of pairs of retinal vessel trees and corresponding retinal images, applying current generative adversarial models. In addition, the dimension of the produced images was $512\times512$, which is greater than commonly generated images on general computer vision problems. We believe that achieving this resolution was only possible due to the constrained class of images in which the method was applied: contrarily to generic natural images, retinal images show a repetitive geometry, where high-level structures such as the field of view, the optical disc, or the macula, are usually present in the image, and act as a guide for the model to learn how to produce new texture and background intensities. 

The main limitation of the presented method is its dependence on a pre-existing vessel tree in order to generate a new image. Furthermore, if the vessel tree comes from the application of a segmentation technique to the original image, the potential weaknesses of the segmentation algorithm will be inherited by the synthesized image. We are currently working on overcoming these challenges. 

\subsubsection*{Acknowledgments}

This work is financed by the ERDF – European Regional Development Fund through the Operational Programme for Competitiveness and Internationalisation - COMPETE 2020 Programme, by National Funds through the FCT – Funda\c{c}\~{a}o para a Ci\^{e}ncia e a Tecnologia (Portuguese Foundation for Science and Technology) within project CMUP-ERI/TIC/0028/2014 and by the North Portugal Regional Operational Programme (NORTE 2020), under the PORTUGAL 2020 Partnership Agreement within the project "NanoSTIMA: Macro-to-Nano Human Sensing: Towards Integrated Multimodal Health Monitoring and Analytics/NORTE-01-0145-FEDER-000016". MDA is the recipient of the Robert C. Watzke Professor of Ophthalmology and Visual Sciences. IDx LLC has no interest in any of the algorithms discussed in this study.

\appendix
\section{Synthetic Retinal Image Quality Evaluation - Discussion}\label{sec_quality_disc}

\subsection{Image Quality Metrics}\label{sec_quality_m}
We discuss now the technical details of the two retinal image quality metrics employed in this work. Regarding the $Q_v$ score \cite{koler_2013}, it is a no-reference quality metric that proceeds by computing a local degree of vesselness around each pixel. This is achieved by building a multiscale version of the input image, represented by the local Hessian matrix around each pixel extracted from the green channel. Frangi's vesselness measure is then computed \cite{frangi_multiscale_1998}, and used as an estimate of visible vessel pixels. Following, an anisotropy measure based on a local Singular Value Decomposition is computed \cite{zhu_automatic_2010}, and the final quality score is obtained as a weighted average of the vesselness map and the local anisotropy values. This way, only vessel pixels are considered in this metric, since these are expected to be good candidates for a reliable contrast and focus estimate.

On the other hand, the Image Structure Clustering (ISC) proposed in \cite{niemeijer_image_2006} follows a substantially different approach. Even if it is also a no-reference quality metric, it is trained on a dataset of retinal images. This dataset contained $1000$ images (independent of our training set) that had been previously labeled by medical experts, depending on whether they showed enough visibility to perform diagnosis. The ISC metric assesses a correct distribution of pixel intensities corresponding to the relevant anatomical structures present in the retina. This is achieved by extracting features consisting of intensities and Gaussian derivatives of the $R$, $G$, and $B$ channels, and then employing k-means to group them into $5$ different clusters. These are observed to be sufficient to model the relevant regions of a retinal image (vessels, optical disk, macula, background-to-foreground and foreground-to-background transitions). Histograms of counts of the computed features are then passed to an SVM, which is trained to predict if the presence and proportion of pixels associated to those structures is consistent, according to the training set correspondent quantities. 

Both metrics seem thus quite complementary, since the ISC technique considers regions from the image that are not addressed by the $Q_v$ score. In our experiments, however, we noticed that the artifacts produced when the generative model was provided an undercomplete vessel tree tended to  rise the ISC score. This drawback was not observed when the $Q_v$ score was computed. 

\begin{figure*}[h]
\centering
\subfloat[$Q_v=\re{0.482}/\syn{0.053}$, ISC = $\re{0.000}/\syn{0.517}$]{\includegraphics[width = 0.48\textwidth]{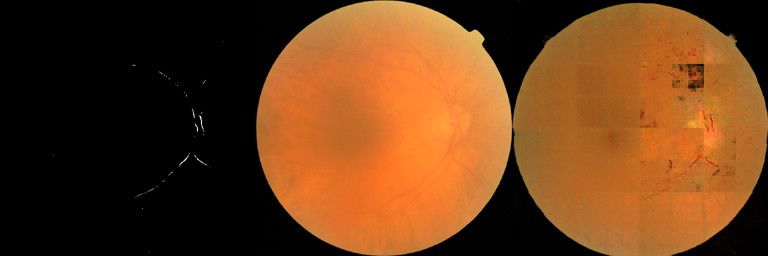}
\label{fig_artifacts_1}}
\hfil
\subfloat[$Q_v=\re{0.115}/\syn{0.097}$, ISC = $\re{0.707}/\syn{0.703}$]{\includegraphics[width = 0.48\textwidth]{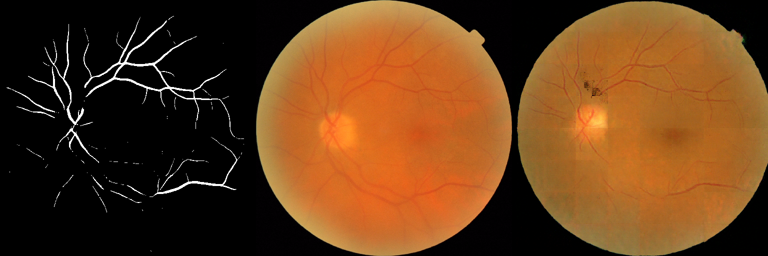}
\label{fig_artifacts_2}}
\caption{A couple of pathological results, in which the generated artifacts artificially rised the ISC quality metric in favor of the synthetic images.}
\label{fig_artifacts}
\end{figure*}

We believe that the reason for this was the following: starting from a real synthetic image, our method employs the vessel tree extracted from it to synthesize a new image; thus, the amount of vessel pixels present in a real image will always be greater than in the corresponding synthetic image, favoring the $Q_v$ score. The ISC metric does not only rely on vessels, but on other anatomical structures. In addition, it considers the three color channels, while the $Q_v$ score employs only one of them. When supplied an image with artifacts such as those in Figure \ref{fig_artifacts}, the ISC score finds that the proportion of colors and edges is not adequate, but still relatively acceptable (note that the scores assigned to the synthetic images are not high in these cases). This situation was detected only on $6$ images from the entire $177$ images present in our test set. Accordingly, for a fair comparison, those images were removed from the statistical experiments that involved the ISC score. Since the $Q_v$ score  seemed to be unaffected by this problem, we include every test image on its analysis.

We believe that current retinal image quality metrics are reasonably suitable to assess the visual quality of synthetic images. However, the study of the anatomical plausibility of these images may benefit of specifically designed quality metrics, that may involve different aspects (local and global) of existing quality assessment approaches.

%
%
\subsection{Supplementary Results}\label{supp_res}
Below we show a random sample of the results produced by our model, together with their real counterparts.

\begin{figure*}[h]
\centering
\subfloat[$Q_v=\re{0.081}/\syn{0.087}$, ISC = $\re{0.993}/\syn{1.0}$]{\includegraphics[width = 0.48\textwidth]{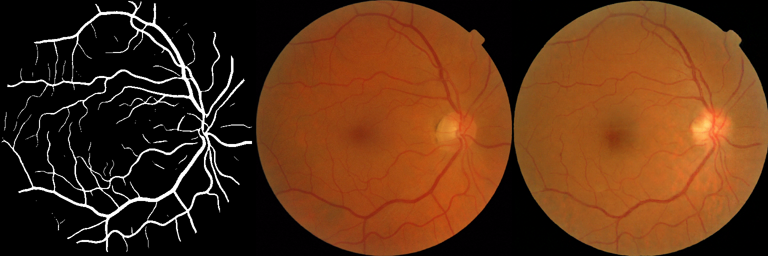}
\label{fig_ret_1}}
\hfil
\subfloat[$Q_v=\re{0.133}/\syn{0.109}$, ISC = $\re{0.983}/\syn{1.0}$]{\includegraphics[width = 0.48\textwidth]{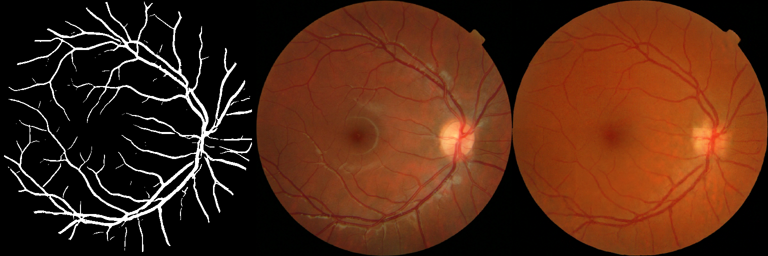}
\label{fig_ret_2}}

\subfloat[$Q_v=\re{0.111}/\syn{0.102}$, ISC = $\re{0.96}/\syn{0.983}$]{\includegraphics[width = 0.48\textwidth]{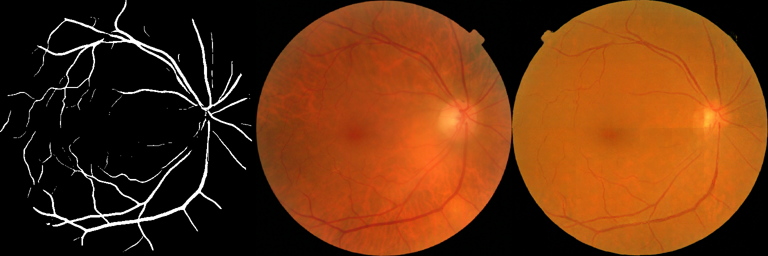}
\label{fig_ret_3}}
\hfil
\subfloat[$Q_v=\re{0.114}/\syn{0.116}$, ISC = $\re{1.0}/\syn{1.0}$]{\includegraphics[width = 0.48\textwidth]{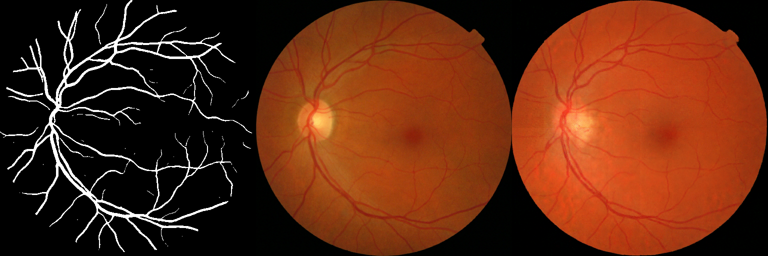}
\label{fig_ret_4}}

\subfloat[$Q_v=\re{0.121}/\syn{0.103}$, ISC = $\re{1.0}/\syn{1.0}$]{\includegraphics[width = 0.48\textwidth]{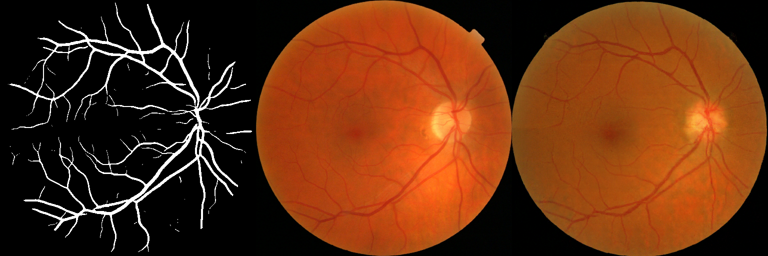}
\label{fig_ret_5}}
\hfil
\subfloat[$Q_v=\re{0.124}/\syn{0.098}$, ISC = $\re{1.0}/\syn{0.996}$]{\includegraphics[width = 0.48\textwidth]{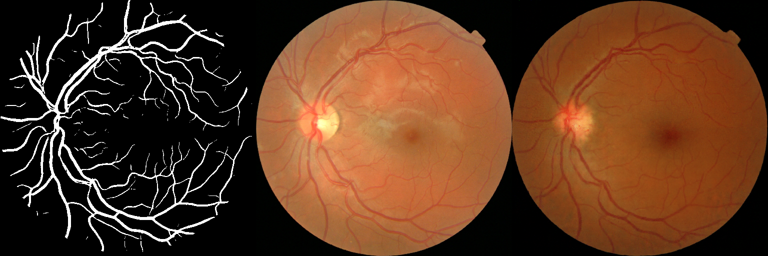}
\label{fig_ret_6}}

\subfloat[$Q_v=\re{0.126}/\syn{0.116}$, ISC = $\re{1.0}/\syn{0.993}$]{\includegraphics[width = 0.48\textwidth]{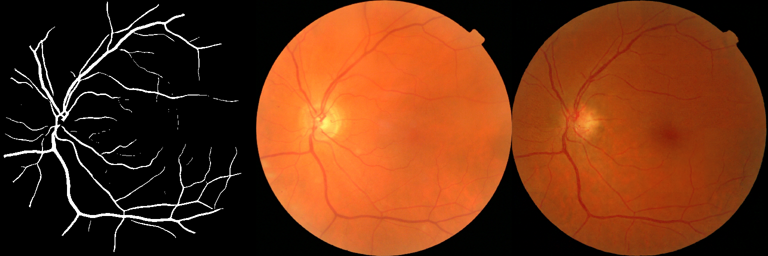}
\label{fig_ret_7}}
\hfil
\subfloat[$Q_v=\re{0.126}/\syn{0.099}$, ISC = $\re{1.0}/\syn{0.996}$]{\includegraphics[width = 0.48\textwidth]{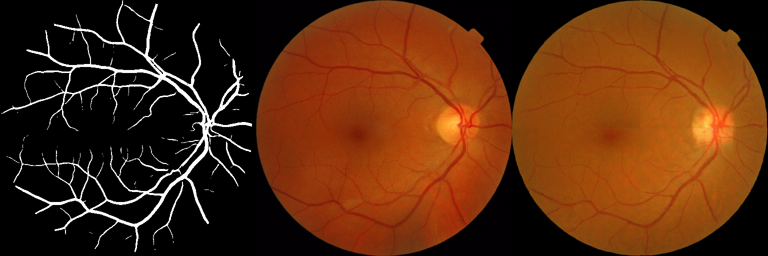}
\label{fig_ret_8}}

\subfloat[$Q_v=\re{0.133}/\syn{0.084}$, ISC = $\re{1.0}/\syn{1.0}$]{\includegraphics[width = 0.48\textwidth]{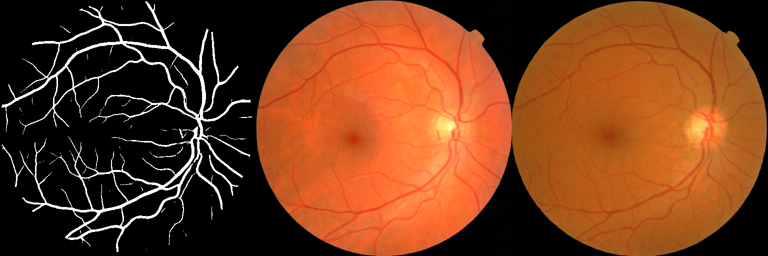}
\label{fig_ret_9}}
\hfil
\subfloat[$Q_v=\re{0.152}/\syn{0.093}$, ISC = $\re{1.0}/\syn{1.0}$]{\includegraphics[width = 0.48\textwidth]{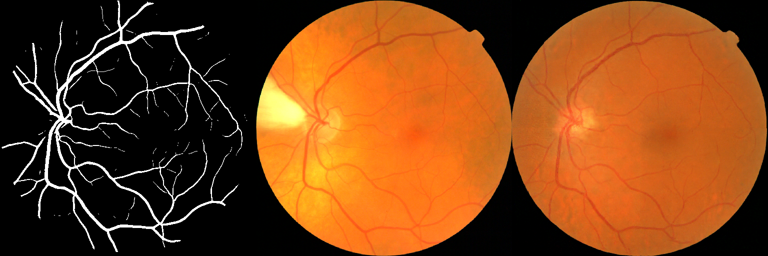}
\label{fig_ret_10}}

\subfloat[$Q_v=\re{0.119}/\syn{0.099}$, ISC = $\re{1.0}/\syn{0.996}$]{\includegraphics[width = 0.48\textwidth]{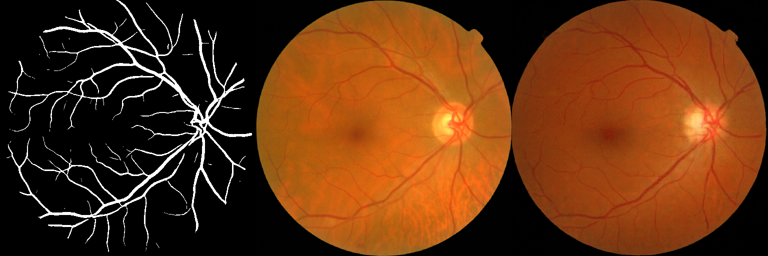}
\label{fig_ret_11}}
\hfil
\subfloat[$Q_v=\re{0.126}/\syn{0.129}$, ISC = $\re{1.0}/\syn{0.99}$]{\includegraphics[width = 0.48\textwidth]{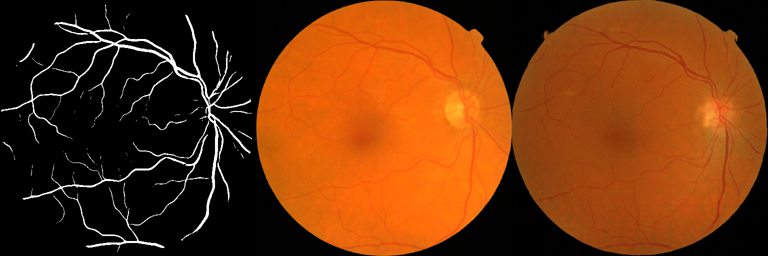}
\label{fig_ret_12}}
\caption{A subset of the generated images. For each block, left: the segmented vessel tree; center: the corresponding real image; right: the synthetic image. Below we show the $Q_v$ and ISC scores for the real (green) and synthetic (red) images. All images are of resolution $512\times512$.}
\label{fig_supp_results}
\end{figure*}

\newpage

Further results are displayed below:

\begin{figure*}[h]
\centering
\subfloat[$Q_v=\re{0.136}/\syn{0.091}$, ISC = $\re{1.0}/\syn{1.0}$]{\includegraphics[width = 0.48\textwidth]{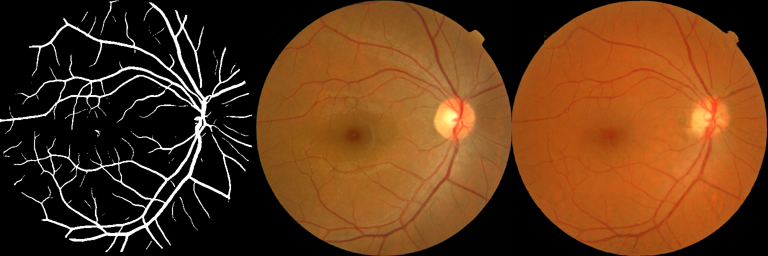}
\label{fig_ret_13}}
\hfil
\subfloat[$Q_v=\re{0.137}/\syn{0.086}$, ISC = $\re{1.0}/\syn{1.0}$]{\includegraphics[width = 0.48\textwidth]{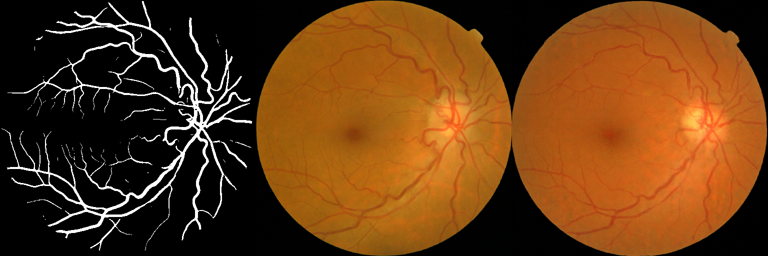}
\label{fig_ret_14}}

\subfloat[$Q_v=\re{0.150}/\syn{0.112}$, ISC = $\re{1.0}/\syn{1.0}$]{\includegraphics[width = 0.48\textwidth]{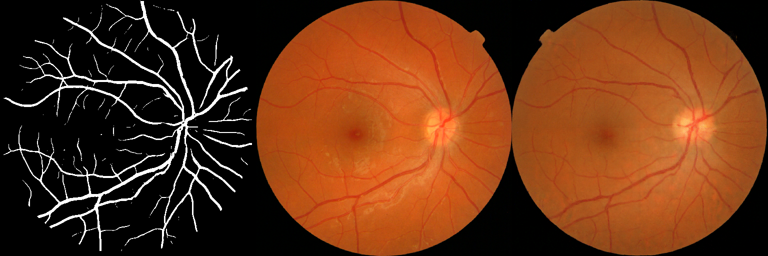}
\label{fig_ret_15}}
\hfil
\subfloat[$Q_v=\re{0.139}/\syn{0.090}$, ISC = $\re{1.0}/\syn{1.0}$]{\includegraphics[width = 0.48\textwidth]{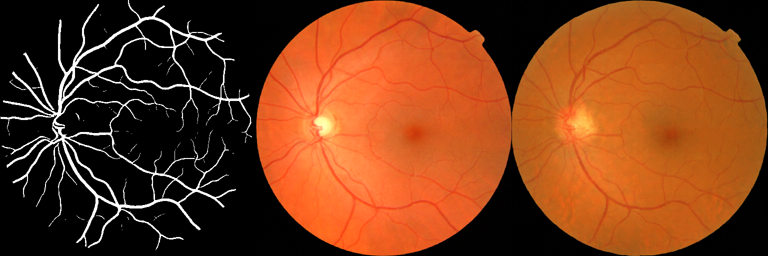}
\label{fig_ret_16}}

\subfloat[$Q_v=\re{0.129}/\syn{0.096}$, ISC = $\re{1.0}/\syn{1.0}$]{\includegraphics[width = 0.48\textwidth]{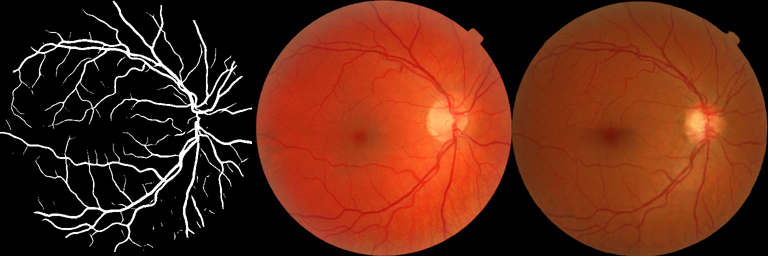}
\label{fig_ret_19}}
\hfil
\subfloat[$Q_v=\re{0.101}/\syn{0.099}$, ISC = $\re{0.99}/\syn{1.0}$]{\includegraphics[width = 0.48\textwidth]{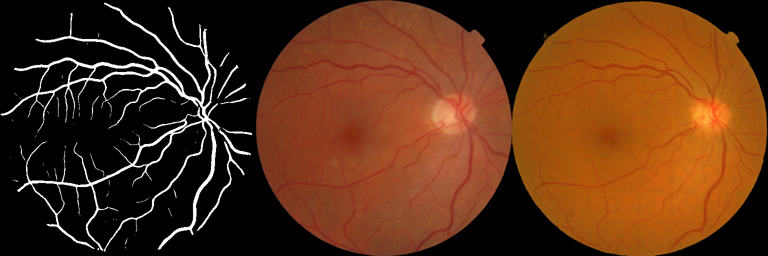}
\label{fig_ret_20}}

\subfloat[$Q_v=\re{0.112}/\syn{0.114}$, ISC = $\re{0.92}/\syn{0.99}$]{\includegraphics[width = 0.48\textwidth]{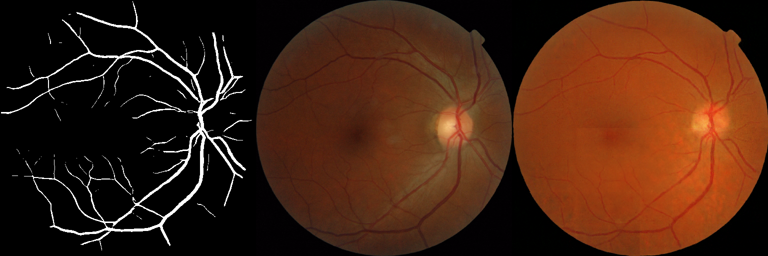}
\label{fig_ret_21}}
\hfil
\subfloat[$Q_v=\re{0.116}/\syn{0.094}$, ISC = $\re{1.0}/\syn{1.0}$]{\includegraphics[width = 0.48\textwidth]{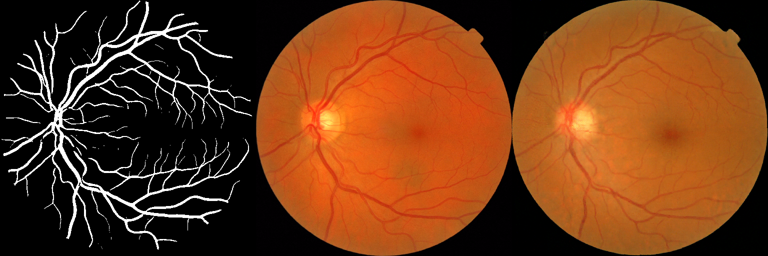}
\label{fig_ret_22}}

\subfloat[$Q_v=\re{0.073}/\syn{0.117}$, ISC = $\re{0.996}/\syn{0.993}$]{\includegraphics[width = 0.48\textwidth]{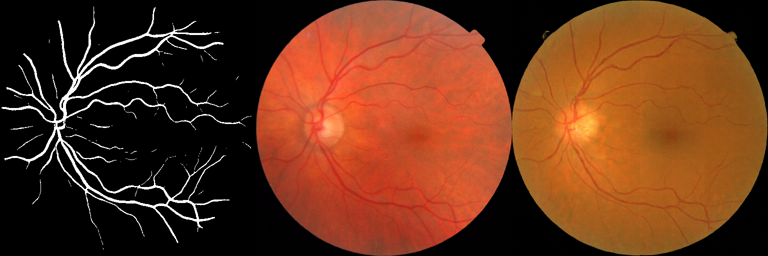}
\label{fig_ret_23}}
\hfil
\subfloat[$Q_v=\re{0.162}/\syn{0.106}$, ISC = $\re{1.0}/\syn{1.0}$]{\includegraphics[width = 0.48\textwidth]{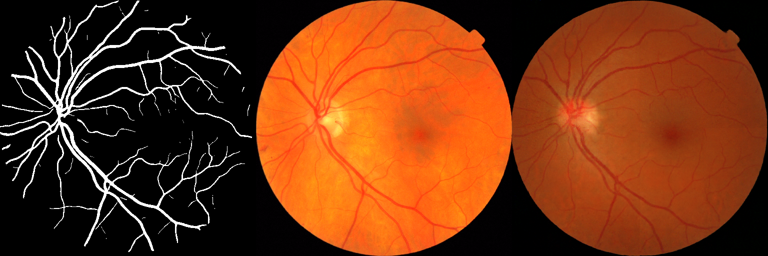}
\label{fig_ret_24}}

\subfloat[$Q_v=\re{0.107}/\syn{0.088}$, ISC = $\re{0.95}/\syn{0.96}$]{\includegraphics[width = 0.48\textwidth]{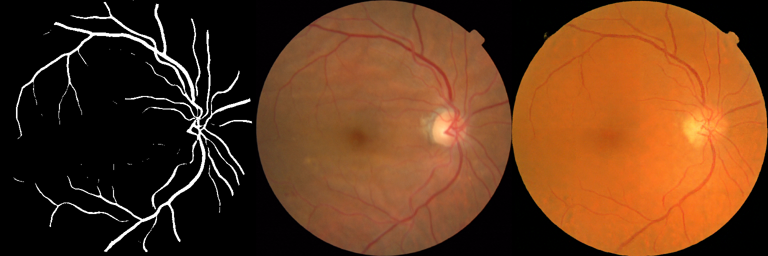}
\label{fig_ret_17}}
\hfil
\subfloat[$Q_v=\re{0.133}/\syn{0.090}$, ISC = $\re{0.983}/\syn{1.0}$]{\includegraphics[width = 0.48\textwidth]{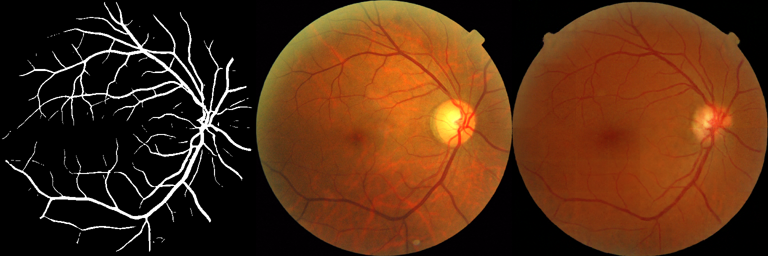}
\label{fig_ret_18}}
\caption{A subset of the generated images. For each block, left: the segmented vessel tree; center: the corresponding real image; right: the synthetic image. Below we show the $Q_v$ and ISC scores for the real (green) and synthetic (red) images. All images are of resolution $512\times512$.}
\label{fig_supp_results2}
\end{figure*}

\newpage

\bibliographystyle{unsrt}
\bibliography{iciar_aris_refs}

\begin{thebibliography}{10}

\bibitem{collins_design_1998}
D.~L. Collins, A.~P. Zijdenbos, V.~Kollokian, J.~G. Sled, N.~J. Kabani, C.~J.
  Holmes, and A.~C. Evans.
\newblock Design and construction of a realistic digital brain phantom.
\newblock {\em IEEE Transactions on Medical Imaging}, 17(3):463--468, June
  1998.

\bibitem{fiorini_automatic_2014}
Samuele Fiorini, Lucia Ballerini, Emanuele Trucco, and Alfredo Ruggeri.
\newblock Automatic generation of synthetic retinal fundus images.
\newblock In Constantino~Carlos Reyes-Aldasoro and Greg Slabaugh, editors, {\em
  Medical {Image} {Understanding} and {Analysis} 2014}, pages 7--12. BMVA
  Press, 2014.

\bibitem{hodneland_physical_2016}
Erlend Hodneland, Erik Hanson, Antonella~Z. Munthe-Kaas, Arvid Lundervold, and
  Jan~M. Nordbotten.
\newblock Physical {Models} for {Simulation} and {Reconstruction} of {Human}
  {Tissue} {Deformation} {Fields} in {Dynamic} {MRI}.
\newblock {\em IEEE Transactions on Bio-Medical Engineering},
  63(10):2200--2210, October 2016.

\bibitem{liu_simulation_2010}
X.~Liu, H.~Liu, A.~Hao, and Q.~Zhao.
\newblock Simulation of {Blood} {Vessels} for {Surgery} {Simulators}.
\newblock In {\em 2010 {International} {Conference} on {Machine} {Vision} and
  {Human}-machine {Interface}}, pages 377--380, April 2010.

\bibitem{cai_integrated_2014}
Jing Cai, You Zhang, Irina Vergalasova, Fan Zhang, W.~Paul Segars, and
  Fang-Fang Yin.
\newblock An {Integrated} {Simulation} {System} {Based} on {Digital} {Human}
  {Phantom} for 4d {Radiation} {Therapy} of {Lung} {Cancer}.
\newblock {\em Journal of Cancer Therapy}, 2014, July 2014.

\bibitem{tulder_why_2015}
Gijs~van Tulder and Marleen~de Bruijne.
\newblock Why {Does} {Synthesized} {Data} {Improve} {Multi}-sequence
  {Classification}?
\newblock In {\em Medical {Image} {Computing} and {Computer}-{Assisted}
  {Intervention} – {MICCAI} 2015}, pages 531--538. Springer International
  Publishing, October 2015.

\bibitem{goodfellow_generative_2014}
Ian Goodfellow, Jean Pouget-Abadie, Mehdi Mirza, Bing Xu, David Warde-Farley,
  Sherjil Ozair, Aaron Courville, and Yoshua Bengio.
\newblock Generative adversarial nets.
\newblock In {\em Advances in {Neural} {Information} {Processing} {Systems}},
  pages 2672--2680, 2014.

\bibitem{isola_image--image_2016}
Phillip Isola, Jun-Yan Zhu, Tinghui Zhou, and Alexei~A. Efros.
\newblock Image-to-{Image} {Translation} with {Conditional} {Adversarial}
  {Networks}.
\newblock {\em arXiv.org}, November 2016.
\newblock arXiv: 1611.07004.

\bibitem{lotter_pgn_2015}
William Lotter, Gabriel Kreiman, and David Cox.
\newblock Unsupervised {Learning} of {Visual} {Structure} using {Predictive}
  {Generative} {Networks}.
\newblock {\em arXiv preprint arXiv:1511.06380}, 2015.

\bibitem{shrivastava_su_learning_2016}
Ashish Shrivastava, Tomas Pfister, Oncel Tuzel, Josh Susskind, Wenda Wang, and
  Russ Webb.
\newblock Learning from {Simulated} and {Unsupervised} {Images} through
  {Adversarial} {Training}.
\newblock {\em arXiv preprint arXiv:1612.07828}, 2016.

\bibitem{ronneberger_u-net:_2015}
Olaf Ronneberger, Philipp Fischer, and Thomas Brox.
\newblock U-{Net}: {Convolutional} {Networks} for {Biomedical} {Image}
  {Segmentation}.
\newblock In {\em Medical {Image} {Computing} and {Computer}-{Assisted}
  {Intervention} – {MICCAI} 2015}, pages 234--241. Springer, Cham, October
  2015.

\bibitem{shelhamer_fcn_2015}
Evan Shelhamer, Jonathan Long, and Trevor Darrell.
\newblock Fully convolutional networks for semantic segmentation.
\newblock {\em Proceedings of the IEEE Conference on Computer Vision and
  Pattern Recognition}, pages 3431--3440, 2015.

\bibitem{staal:2004-855}
J.~J. Staal, M.~D. Abramoff, M.~Niemeijer, M.~A. Viergever, and B.~van
  Ginneken.
\newblock Ridge based vessel segmentation in color images of the retina.
\newblock {\em IEEE Transactions on Medical Imaging}, 23(4):501--509, 2004.

\bibitem{kingma:adam_2014}
Diederik Kingma and Jimmy Ba.
\newblock Adam: {A} {Method} for {Stochastic} {Optimization}.
\newblock {\em International Conference on Learning Representations}, pages
  1--13, 2014.

\bibitem{Liskowski_2016}
P.~Liskowski and K.~Krawiec.
\newblock Segmenting retinal blood vessels with deep neural networks.
\newblock {\em IEEE Transactions on Medical Imaging}, 35(11):2369--2380, Nov
  2016.

\bibitem{Youden_1950}
W.~J. Youden.
\newblock Index for rating diagnostic tests.
\newblock {\em Cancer}, 3(1):32--35, 1950.

\bibitem{decenciere_feedback_2014}
Etienne Decencière, Xiwei Zhang, Guy Cazuguel, Bruno Lay, Béatrice Cochener,
  Caroline Trone, Philippe Gain, Richard Ordonez, Pascale Massin, Ali Erginay,
  Béatrice Charton, and Jean-Claude Klein.
\newblock Feedback on a publicly distributed database: the {Messidor} database.
\newblock {\em Image Analysis \& Stereology}, 33(3):231--234, August 2014.

\bibitem{chollet_keras_2015}
Fran\c{c}ois Chollet.
\newblock Keras.
\newblock \url{https://github.com/fchollet/keras}, 2015.

\bibitem{wang_why_2002}
Z.~Wang, A.~C. Bovik, and L.~Lu.
\newblock Why is image quality assessment so difficult?
\newblock In {\em 2002 {IEEE} {International} {Conference} on {Acoustics},
  {Speech}, and {Signal} {Processing}}, volume~4, pages IV--3313--IV--3316, May
  2002.

\bibitem{theis_note_2016}
L.~Theis, A.~van~den Oord, and M.~Bethge.
\newblock A note on the evaluation of generative models.
\newblock In {\em International {Conference} on {Learning} {Representations}},
  2016.

\bibitem{koler_2013}
T.~Köhler, A.~Budai, M.~F. Kraus, J.~Odstrčilik, G.~Michelson, and
  J.~Hornegger.
\newblock Automatic no-reference quality assessment for retinal fundus images
  using vessel segmentation.
\newblock In {\em Proceedings of the 26th IEEE International Symposium on
  Computer-Based Medical Systems}, pages 95--100, June 2013.

\bibitem{niemeijer_image_2006}
Meindert Niemeijer, Michael~D. Abràmoff, and Bram van Ginneken.
\newblock Image structure clustering for image quality verification of color
  retina images in diabetic retinopathy screening.
\newblock {\em Medical Image Analysis}, 10(6):888--898, December 2006.

\bibitem{frangi_multiscale_1998}
Alejandro~F. Frangi, Wiro~J. Niessen, Koen~L. Vincken, and Max~A. Viergever.
\newblock Multiscale vessel enhancement filtering.
\newblock In {\em Medical {Image} {Computing} and {Computer}-{Assisted}
  {Intervention} — {MICCAI}’98}, pages 130--137. Springer, Berlin,
  Heidelberg, October 1998.
\newblock DOI: 10.1007/BFb0056195.

\bibitem{zhu_automatic_2010}
X.~Zhu and P.~Milanfar.
\newblock Automatic {Parameter} {Selection} for {Denoising} {Algorithms}
  {Using} a {No}-{Reference} {Measure} of {Image} {Content}.
\newblock {\em IEEE Transactions on Image Processing}, 19(12):3116--3132,
  December 2010.

\end{thebibliography}

\end{document}